\documentclass[10pt,conference,a4paper]{IEEEtran}

\usepackage{graphicx}
\usepackage{multirow}
\usepackage{microtype}
\usepackage{booktabs}
\usepackage[hidelinks]{hyperref}

\emergencystretch =\maxdimen
\begin{document}

\title{Quantifying Radiographic Knee Osteoarthritis Severity using Deep Convolutional Neural Networks} 

\author{
    \IEEEauthorblockN{Joseph Antony\IEEEauthorrefmark{1}, Kevin McGuinness\IEEEauthorrefmark{1}, Noel E O'Connor\IEEEauthorrefmark{1}, Kieran Moran\IEEEauthorrefmark{1}\IEEEauthorrefmark{2}}
    
    \IEEEauthorblockA{\IEEEauthorrefmark{1}Insight Centre for Data Analytics, Dublin City University }
    
    \IEEEauthorblockA{\IEEEauthorrefmark{2}School of Health and Human Performance, Dublin City University}
\ joseph.antony2@mail.dcu.ie
}

\maketitle

\begin{abstract}
This paper proposes a new approach to automatically quantify the severity of knee osteoarthritis (OA) from radiographs using deep convolutional neural networks (CNN). Clinically, knee OA severity is assessed using Kellgren \& Lawrence (KL) grades, a five point scale. Previous work on automatically predicting KL grades from radiograph images were based on training shallow classifiers using a variety of hand engineered features. We demonstrate that classification accuracy can be significantly improved using deep convolutional neural network models pre-trained on ImageNet and fine-tuned on knee OA images. Furthermore, we argue that it is more appropriate to assess the accuracy of automatic knee OA severity predictions using a continuous distance-based evaluation metric like mean squared error than it is to use classification accuracy. This leads to the formulation of the prediction of KL grades as a regression problem and further improves accuracy. Results on a dataset of X-ray images and KL grades from the Osteoarthritis Initiative (OAI) show a sizable improvement over the current state-of-the-art.

\end{abstract}



\section{Introduction}

The increasing prevalence of knee osteoarthritis (OA), a degenerative joint disease, and total joint arthoplasty as a serious consequence, means there is a growing need for effective clinical and scientific tools to diagnose knee OA in the early stage, and to assess its severity in progressive stages~\cite{oka2008, shamir2009}. Detecting knee OA and assessing the severity of knee OA are crucial for pathology, clinical decision making, and predicting disease progression \cite{braun2012}. 
Joint space narrowing (JSN) and osteophytes (bone spurs) formation are the key pathological features of knee OA \cite{oka2008}, which are easily visualized using radiographs~\cite{braun2012}. 
 
 The assessment of knee OA severity has traditionally been approached as an image classification problem~\cite{shamir2009}, with the KL grades being the ground truth for classification. Radiographic features detectable through a computer-aided analysis are clearly useful to quantify knee OA severity, and to predict the future development of knee OA~\cite{shamir2009}. However, based on the results reported, the  accuracy of both the multi-class and consecutive grades classification is far from ideal.  Previous work on classifying knee OA from radiographic images have used Wndchrm, a multipurpose bio-medical image classifier~\cite{shamir2008,orlov2008}. 
 The feature space used by Wndchrm includes hand-crafted features to capture these characteristics based on polynomial decomposition, contrast, pixel statistics, textures and also features extracted from image transforms~\cite{shamir2009,shamir2008,orlov2008}. 
 
Instead of hand-crafted features, we propose that learning feature representations using a CNN can be more effective for classifying knee OA images to assess the severity condition. Feature learning approaches provide a natural way to capture cues by using a large number of code words (sparse coding) or neurons (deep networks), while traditional computer vision features, designed for basic-level category recognition, may eliminate many useful cues during feature extraction \cite{yang2013}. 
Manually designed or hand-crafted features often simplify machine learning tasks. Nevertheless, they have a few disadvantages. The process of engineering features requires domain-related expert knowledge, and is often very time consuming~\cite{lee2010}. These features are often low-level as prior knowledge is hand-encoded, and features in one domain do not always generalize to other domains~\cite{le2013}. In recent years, learning feature representations is preferred to hand-crafted features, particularly for fine-grained classification, because rich appearance and shape features are essential for describing subtle differences between categories~\cite{yang2013}. 
 
 A convolutional neural network (CNN) typically comprises multiple convolutional and sub-sampling layers, optionally followed by fully-connected layers like a standard multi-layer neural network. A CNN exploits the 2D spatial structure images to learn translation invariant features. This is achieved with local connections and associated weights followed by some form of pooling. The main advantage of CNN over fully-connected networks is that they are easier to train and have fewer parameters with the same number of hidden units~\cite{prasoon2013}.
 
In this work, first, we investigated the use of well-known CNNs such as the VGG 16-layer net~\cite{simonyan2014}, and comparatively simpler networks like VGG-M-128~\cite{chatfield2014}, and BVLC reference CaffeNet~\cite{jia2014,karayev2013} (which is very similar to the widely-used \textit{AlexNet} model~\cite{krizhevsky2012imagenet}) to classify knee OA images. These networks are pre-trained for color image classification using a very large dataset such as the ImageNet LSVRC dataset~\cite{russakovsky2015imagenet}, which contains 1.2 million images with 1000 classes. Initially, we extracted features from the convolutional, pooling, and fully-connected layers of VGG16, VGG-M-128, and BVLC CaffeNet, and trained linear SVMs to classify knee OA images.

Next, motivated by the transfer learning approach~\cite{yosinski2014}, we fine-tuned the pre-trained networks. We adopted transfer learning as the OAI dataset we work with is small, containing only a few thousand images. In this setting, a base network is first trained on external data, and then the weights of the initial $n$ layers are transferred to a target network~\cite{yosinski2014}. The new layers of the target network are randomly initialized. Intuitively, the lower layers of the networks contain more generic features such as edge or texture detectors useful for multiple tasks, while the upper layers progressively focus on more task specific cues~\cite{karayev2013,yosinski2014}. We used this approach for both classification and regression, adding new fully-connected layers and use backpropagation to fine tune the weights for the complete network on the target loss.

The primary contributions of this paper are the use of CNNs and regression loss to quantify knee OA severity. We propose the use of mean squared error for assessing the performance of an automatic knee OA severity assessment instead of binary and multi-class classification accuracy. We show that the inferred CNN features from the fine-tuned BVLC reference CaffeNet provide higher classification accuracy in comparison to the state-of-the-art. We also present an SVM-based method to automatically detect and extract the knee joints from knee OA radiographs.

\section{Materials and Methods}

\subsection{Dataset}
\begin{figure}[t]
  \centering
  \includegraphics[width = 0.45 \textwidth, height = 0.25 \textwidth]{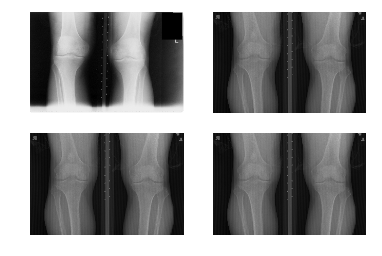}
  \caption{A few samples of bilateral PA fixed flexion knee OA radiographs.}
  \label{fig:sam}
  \vspace{-0.4 cm}
\end{figure}

\begin{figure}[t]
  \centering
  \includegraphics[width = 0.45 \textwidth, height = 0.25 \textwidth]{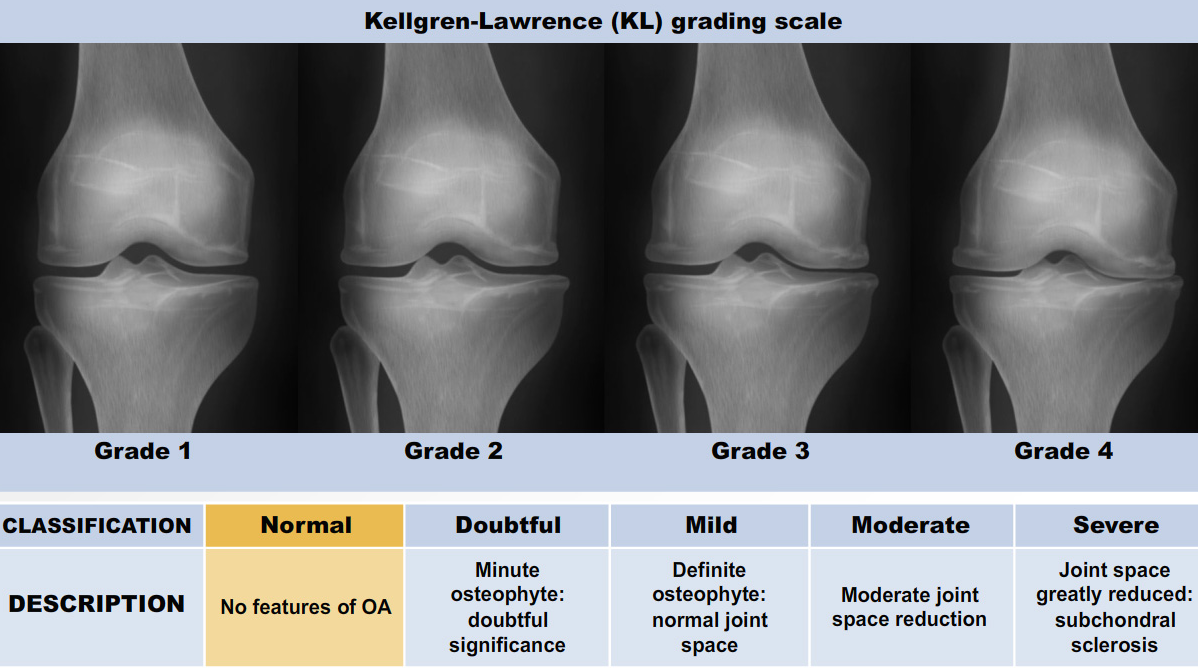}
  \caption{The KL grading system to assess the severity of knee OA.
  	\footnotesize{Source: \url{http://www.adamondemand.com/clinical-management-of-osteoarthritis/}}
  }
  \label{fig:KL}
\end{figure}

The data used for the experiments are bilateral PA fixed flexion knee X-ray images, taken from the baseline (image release version O.E.1) radiographs of the Osteoarthritis Initiative (OAI) dataset containing an entire cohort of $4,476$ participants. This is a standard dataset for studies involving knee OA. Figure~\ref{fig:sam} shows some samples from the dataset. In the entire cohort, Kellgren \& Lawrence (KL) grades are available for both knee joints in $4,446$ radiographs and these images were used for this study. The  distribution of the knee joint images (in total $8,892$) conditioned on the KL grading scale are: Grade 0 - 3433, Grade 1 - 1589, Grade 2 - 2353, Grade 3 - 1222, and Grade 4 - 295. The KL grading system uses 5 grades to classify knee OA severity from the radiographs \cite{park2013}, where `Grade 0' corresponds to the normal knee, and the other grades correspond to the progression of the disease, as shown in Figure~\ref{fig:KL}.

\subsection{Automatic detection and the extraction of the knee joints}
Automatically detecting, and extracting the knee joint region from the radiographs is an important pre-processing step and Shamir et. al.~\cite{shamir2009} proposed the template matching method for this. Though this method is simple to implement, the accuracy of detecting the knee joints is low for our dataset. To improve detection, we propose an SVM-basd method.

\subsubsection{Template matching}
As a baseline, we adapted the template matching approach~\cite{shamir2009} for detecting the knee joint center, to an image patch of size 20$\times$20 pixels. The radiographs are first down-scaled to 10\% of the original size and subjected to histogram equalization for intensity normalization. An image patch (20$\times$20 pixels) containing the knee joint center is taken as a template. 10 image patches from each grade, so that in total 50 patches were pre-selected as templates.  Each input image is scanned by an overlapping sliding window (20$\times$20 pixels). At each window the Euclidean distance between the image patch and the 50 templates are calculated, and the shortest distance is recorded. After scanning an entire image with the sliding window, the window that records the smallest Euclidean distance is recorded as the knee joint center.

\subsubsection{Proposed method for detecting the knee joints}
\begin{figure}[t]
  \centering
  \includegraphics[width = 0.4 \textwidth, height = 0.2 \textwidth]{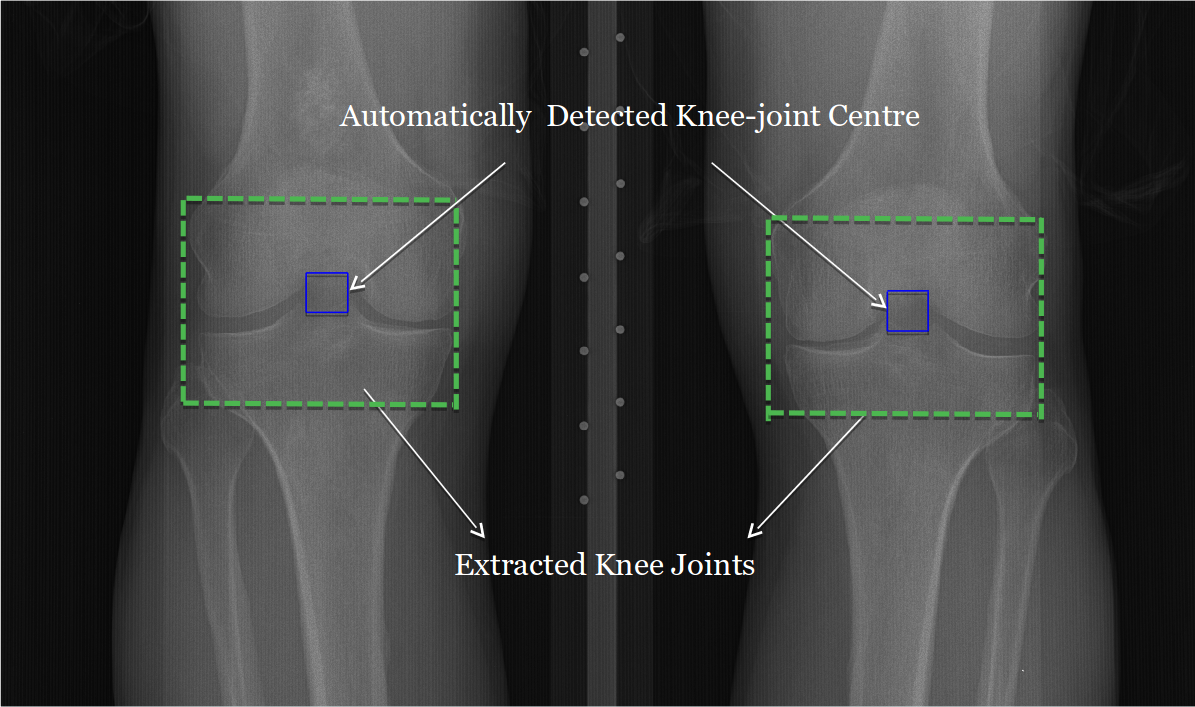}
  \caption{Detecting the knee joint centers and extracting the knee joints.}
  \label{fig:AutoDet}
  \vspace{-0.4 cm}
\end{figure}

We propose an approach using a linear SVM and the Sobel horizontal image gradients as the features for detecting the knee joint centers. The well-known Sobel edge detection algorithm uses the vertical and the horizontal image gradients. The  motivation for this is that knee joint images primarily contain horizontal edges. The image patches (20$\times$20 pixels) containing the knee joint center are taken as the positive training samples and the image patches (20$\times$20 pixels) excluding the knee joint center are taken as the negative training samples. After extracting Sobel horizontal gradients for the positive and negative samples, a linear SVM was trained. To detect the knee joint center from both left and right knees, input images are split in half to isolate left and right knees separately. A sliding window (20$\times$20 pixels) is used on either half of the image, and the Sobel horizontal gradient features are extracted for every image patch. The image patch with the maximum score based on the SVM decision function is recorded as the detected knee joint center, and the area (300$\times$300 pixels) around the knee joint center is extracted from the input images using the corresponding recorded coordinates. Figure \ref{fig:AutoDet} shows an example of a detected and extracted knee joint.

\subsection{Assessing the knee OA severity using CNNs}
In this study, we investigate the use of CNN for assessing the severity of knee OA through classification and regression. For this, we used two approaches: 1. Pre-trained CNN for fixed feature extraction, 2. Fine-tuning the pre-trained CNN following the transfer learning approach. For benchmarking the classification results obtained by the proposed methods, we have used Wndchrm, an open source utility for medical image classification that has been applied to this task in the literature~\cite{shamir2008,shamir2009}.

\subsubsection{Classification using features extracted from pre-trained CNNs}
As our initial approach, we trained VGG16~\cite{simonyan2014} with the OAI dataset. We used the Caffe~\cite{jia2014} framework for implementing and training the CNN, and to extract features from the CNN. We extracted features from the different layers of the VGG net such as fully-connected (fc7), pooling (pool5), and convolutional (conv5\_2) layers to identify the most discriminating set of features. Linear SVMs (trained using LIBLINEAR~\cite{fan2008liblinear}) were trained with the extracted CNN features for classifying knee OA images, where the ground truth was labeled images conditioned on the KL grades. Next, we investigated the use of simpler pre-trained CNNs such as VGG-M-128~\cite{chatfield2014} and BVLC CaffeNet~\cite{jia2014} for classifying the knee OA images. These networks have fewer layers and parameters in comparison to VGG16. 
  
\subsubsection{Fine-tuning the CNNs for classification and regression}
Our next approach fine-tuned the BVLC CaffeNet~\cite{jia2014} and VGG-M-128~\cite{chatfield2014} networks. We chose these two smaller networks, both which contain fewer layers and parameters ($\sim$62M), over the much deeper VGG16, which has $\sim$138M parameters. We replace the top fully-connected layer of both networks and retrain the model on the OAI dataset using backpropagation. The lower-level features in the bottom layers are also updated during fine-tuning. Standard softmax loss was used as the objective for classification, and accuracy layers were added to monitor training progress. A Euclidean loss layer (mean squared error) was used for the regression experiments.


\section{Results and Discussion}

\subsection{Automatic detection of the knee joints}
Standard template matching~\cite{shamir2009} produces poor detection accuracy on our dataset. To improve this, we used a linear SVM with the Sobel horizontal image gradients as features to detect the knee joints. The proposed method is approximately $80\times$ faster than template matching; for detecting all the knee joints in the dataset comprising $4,492$ radiographs, the proposed method took $\sim$9 minutes and the template matching method took $\sim$798 minutes.

Image patches containing the knee joint center (20$\times$20 pixels) were used as positive examples and randomly sampled patches excluding the knee joint as negative samples. We used 200 positive and 600 negative training samples. The samples were split into 70\% training and 30\% test set. Fitting a linear SVM produced $\textbf{95.2\%}$ 5-fold cross validation and $\textbf{94.2\%}$ test accuracies. Table~\ref{Tab:AD} shows the precision, recall, and $F_{1}$scores of this classification.


\begin{table}[t]
\caption{Classification metrics of the SVM for detection.}
\label{Tab:AD}
\centering
\begin{tabular}{ c  c  c  c  c }
\toprule
Class & Precision & Recall & $F_{1}$score\\
\midrule
 Positive & 0.93 & 0.84 & 0.88  \\
 Negative & 0.95 & 0.98 & 0.96  \\
\midrule
 Mean & 0.94 & 0.94 & 0.94\\
 \bottomrule
\end{tabular}
\end{table}

To evaluate the automatic detection, we generated the ground truth by manually annotating the knee joint centers (20$\times$20 pixels) in 4,496 radiographs using an annotation tool that we developed, which recorded the bounding box (20$\times$20 pixels) coordinates of each annotation.

We use the well-known Jaccard index to give a matching score for each detected instance. The Jaccard index J(A,D) is given by,
\begin{equation}
J(A,D) = \frac{A \cap D} {A \cup D}
\end{equation}
where A, is the manually annotated and D is the automatically detected knee joint center using the proposed method. Table~\ref{Tab:Jac} shows the resulting average detection accuracies based on thresholding of Jaccard indices.

\begin{table}[t]
\caption{Comparison of automatic detection using the template matching and the proposed method based on Jaccard Index (J).}
\label{Tab:Jac}
\centering
\begin{tabular}{l c c c}
\toprule
Method & $J=1$ & $J\geq0.5$ & $J>0$\\
\midrule
 Template Matching & 0.3 \% & 8.3 \% & 54.4 \% \\
 Proposed Method & 1.1 \% & 38.6 \% & \textbf{81.8 \%} \\
\bottomrule
\end{tabular}
\end{table}

\begin{table*}[!t]
\caption{Classification accuracy (\%) achieved by the Wndchrm and pre-trained CNN features.}
\centering
\begin{tabular}{c c c c c c c c c c c c}
 \toprule
& \multirow{2}{*}{Classification} & \multirow{2}{*}{Wndchrm} & \multicolumn{3}{c}{VGG 16-Layers Net} & \multicolumn{3}{c}{VGG-M-128 Net}&  \multicolumn{3}{c}{BVLC ref CaffeNet}\\ 
\cmidrule{4-12}
&  &  & fc7 & pool5 & conv5\_2 & fc6 & pool5 & conv4 & fc7 & pool5 & conv5\\
 \midrule
\multirow{4}{*}{Progressive} & Grade 0 vs Grade 1 & 51.5 &  56.3 &  61.3 & 63.5 & 56.5 & 63.2 & \textbf{64.7} & 62.0  & 64.3 & 63.3\\  
& Grade 0 vs Grade 2 & 62.6 &  68.6 &  74.3 & 76.7 & 67.8 & 75.5 & \textbf{77.6} & 69.6  & 73.6 & 73.9\\ 
& Grade 0 vs Grade 3 & 70.6 &  86.4 &  91.4 & 92.4 & 88.5 & 90.2 & \textbf{92.9} & 87.9  & 92.5 & 91.5\\ 
& Grade 0 vs Grade 4 & 82.8 &  98.1 &  98.6 & 99.3 & 98.8 & 99.3 & 99.2 & 98.5  & \textbf{99.4} & 99.1\\  
 \midrule
 
 \multirow{3}{*}{Successive} & Grade 1 vs Grade 2 & 48.8 &  60.0 &  64.7 & 67.3 & 57.9 & 63.5 & 65.3 & 61.2  & \textbf{65.8} & 62.8 \\
& Grade 2 vs Grade 3 & 54.5 &  69.8 &  76.4 & 77.0 & 73.0 & 77.3 & \textbf{79.0} & 70.3  & 78.1 & 77.1\\ 
& Grade 3 vs Grade 4 & 58.6 &  85.2 &  88.8 & 90.0 & 85.0 & 90.4 & 91.2 & 87.4 & \textbf{91.6} & 91.4\\ 
\midrule

\multirow{3}{*}{Multi-class} & Grade 0 to Grade 2 & 39.9 &  51.1 &  53.4 & 56.9 & 51.1 & 55.0 & \textbf{57.4} & 51.1  & 54.8 & 54.4\\  
& Grade 0 to Grade 3 & 32.0 &  44.6 &  48.7 & 53.9 & 45.4 & 50.2 & \textbf{53.3} & 46.9  & 51.6 & 50.2\\ 
& Grade 0 to Grade 4 & 28.9 &  42.6 &  47.6 & 53.1 & 43.8 & 49.5 & \textbf{53.4} & 44.1  & 50.8 & 50.0\\ 
\bottomrule
\end{tabular}
\label{Tab:Clsf_PT}
\vspace{-0.4cm}
\end{table*}

The mean Jaccard index for the template matching and the classifier methods are $\textbf{0.1}$ and $\textbf{0.36}$. From Table~\ref{Tab:Jac}, it is evident that the proposed method is more accurate than template matching. This is due to the fact that template matching relies upon the intensity level difference across an input image. Thus, it is prone to matching a patch with small Euclidean distance that does not actually correspond to the joint center. We also varied the templates in a set, and observed that the detection is highly dependent on the choice of templates: template matching is similar to a k-nearest neighbor classifier with $k=1$. The reason for higher accuracy in the proposed method is the use of horizontal edge detection instead of intensity level differences. The knee joints primarily contain horizontal edges and thus are easily detected by the classifier using horizontal image gradients as features.

Despite sizable improvements in accuracy and speed using the proposed approach, detection accuracy still falls short of 100\%. We therefore decided to use our manual annotations so as to investigate KL grade classification performance independently of knee joint detection.

\subsection{Classification of the knee joints using pre-trained CNNs}

The extracted knee joint images were split into training ($\sim$70\%) and test ($\sim$30\%) as per the KL grades. 
For classifying the knee joint images, we extracted features from fully-connected, pooling and convolution layers of VGG16, VGG-M-128, and BVLC CaffeNet. For binary and multi-class classifications, linear SVMs were trained individually with the extracted features. The classification results achieved with the CNNs are compared to knee classification of OA images using the Wndchrm~\cite{shamir2009,shamir2008,orlov2008}. 




Table~\ref{Tab:Clsf_PT} shows the test set classification accuracies achieved by Wndchrm and the CNN features. The CNN features consistently outperform Wndchrm for classifying healthy knee samples against the progressive stages of knee OA. The features from conv4 layer with dimension 512$\times$13$\times$13 and pool5 layer 256$\times$13$\times$13 of VGG-M-128 net, and conv5 layer with dimension 512$\times$6$\times$6 and pool5 layer with dimension 256$\times$6$\times$6 of BVLC reference CaffeNet give higher classification accuracy in comparison to the fully-connected fc6 and fc7 layers of VGG nets and CaffeNet. 
We also extracted features from further bottom layers such as pool4, conv4\_2, pool3, pool2 and trained classifiers on top of these features. As the dimension of the bottom layers are high, significantly more time was required for training but without improvement in  classification accuracy.  


In a fine-grained classification task such as knee OA images classification, the accuracy of classifying successive classes tends to be low, as the variations in the progressive stages of the disease are minimal, and only highly discriminant features can capture these variations. From the experimental results, as shown in Table \ref{Tab:Clsf_PT}, the features extracted from CNNs provide significantly higher classification accuracy in comparison to the Wndchrm, and these features are effective and promising for classifying the consecutive stages of knee OA.


We performed multi-class classifications using linear SVMs with the CNN features (Table~\ref{Tab:Clsf_PT}, multi-class). Again, the CNN features perform significantly better than the Wndchrm-based approach. The classification accuracies obtained using convolutional (conv4, conv5) and pooling (pool5) layers are slightly higher in comparison to fully-connected layer features. There are minimal variations in classification accuracy obtained with the features extracted from VGG-M-128 net and BVLC reference CaffeNet in comparison to VGG16. 

\subsection{Classification of the knee joints using fine-tuned CNNs}

\begin{figure}[b]
  \begin{minipage}[b]{0.5\linewidth}
    \centering
    \includegraphics[width=1\textwidth] {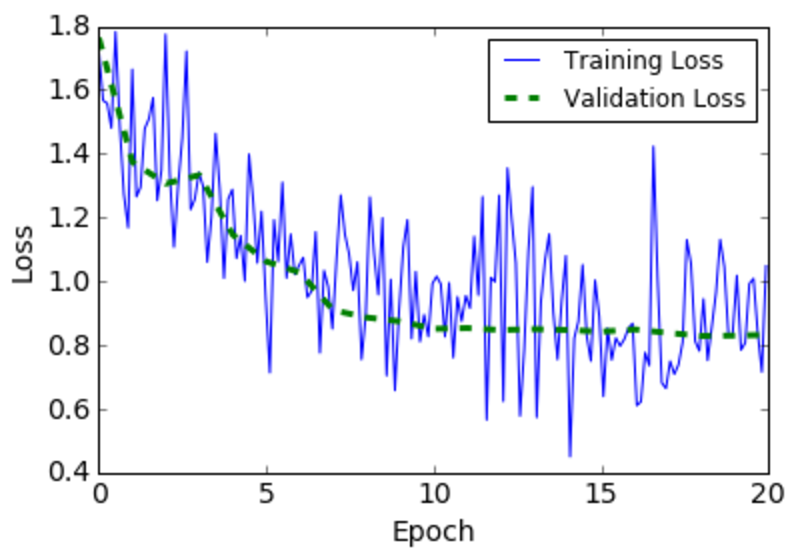}
  \end{minipage}%
  \begin{minipage}[b]{0.5\linewidth}
    \centering
	\includegraphics[width=1\textwidth]{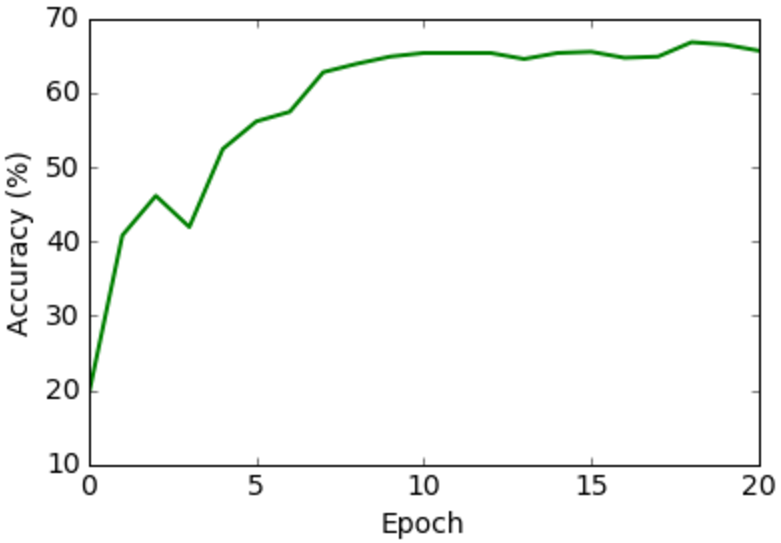}
    \par\vspace{0pt}
\end{minipage}
\caption{Learning curves for training and validation loss (left) and validation accuracy (right) during fine-tuning.}
\label{fig:LossAcc}
\vspace{-0.4cm}
\end{figure}

Table~\ref{Tab:Clsf_PT} shows the multi-class classification results for the fine-tuned BVLC CaffeNet and VGG-M-128 networks. We omitted the VGG16 network in these experiment since the variation in accuracy among the pre-trained CNNs was small, and fine-tuning VGG16 is significantly more computationally expensive.
%
%
The dataset was split into training (60\%), validation (10\%) and test (30\%) sets for fine-tuning. To increase the number of training samples, we included the right-left flipped knee joint images in the training set. The networks were fine-tuned for 20 epochs using a learning rate of 0.001 for the transferred layers, and boosting it on newly introduced layers by a factor of 10. The performance of fine-tuned BVLC CaffeNet was slightly better than VGG-M-128. Hence, we only show here the results of fine-tuning CaffeNet. Figure~\ref{fig:LossAcc} shows the learning curves for training and validation loss, and validation accuracy. The decrease in loss and increase in accuracy shows that the fine-tuning is effective and makes the CNN features more discriminative, which improves classification accuracy (Table~\ref{Tab:Clsf_PT}). The features extracted from the fully connected (fc7) layer provide slightly better classification in comparison to pooling (pool5) and convolution (conv5) layers.




\begin{table}[t]
\caption{Classification accuracy (\%) achieved with the features extracted from fine-tuned BVLC Net.}
\centering
\begin{tabular}{c c c c c c c}
 \toprule
\multirow{2}{*}{Classification} &  \multicolumn{3}{c}{Before Fine-Tuning} & \multicolumn{3}{c}{After Fine-Tuning}\\ 
\cmidrule{2-7}
 & fc7 & pool5 & conv5 & fc7 & pool5 & conv5\\
 \midrule
Grade 0 vs Grade 1 & 62.0 &  64.3 &  63.3 & 63.3 & \textbf{64.3} & 61.9\\  
Grade 0 vs Grade 2 & 69.6 &  73.6 &  73.9 & 76.3 & \textbf{77.2} & 74.1\\ 
Grade 0 vs Grade 3 & 87.9 &  92.5 &  91.5 & \textbf{96.7} & 96.0 & 96.3\\ 
Grade 0 vs Grade 4 & 98.5 &  99.4 &  99.1 & \textbf{99.8} & 99.7 & 99.7\\  
 \midrule
 
Grade 1 vs Grade 2 & 61.2 &  65.8 &  62.8 & 63.3 & \textbf{66.7} & 62.7\\ 
Grade 2 vs Grade 3 & 70.3 &  78.1 &  77.1 & \textbf{85.8} & 83.9 & 83.3\\ 
Grade 3 vs Grade 4 & 87.4 &  91.6 &  91.4 & \textbf{94.4} & 93.6 & 92.6\\  
 \midrule
 
Grade 0 to Grade 2 & 51.1 &  54.8 &  54.4 & \textbf{57.4} & 57.0 & 52.0\\  
Grade 0 to Grade 3 & 46.9 &  51.6 &  50.2 & \textbf{57.2} & 56.5 & 51.8\\ 
Grade 0 to Grade 4 & 44.1 &  50.8 &  50.0 & \textbf{57.6} & 56.2 & 51.8\\ 
\bottomrule

\end{tabular}
\label{Tab:Clsf_FT}
\vspace{-0.4cm}
\end{table}

\subsection{Regression of KL grades using fine-tuned CNNs.}

Existing work on automatic measurement of knee OA severity treats it as an image classification problem, assigning each KL grade to a distinct category \cite{shamir2009}. To date, evaluation of automatic KL grading algorithms has been based on binary and multi-class classification accuracy with respect to these discrete KL grades \cite{oka2008,shamir2009,orlov2008}. KL grades are not, however, categorical, but rather represent an ordinal scale of increasing severity. Treating them as categorical during evaluation means that the penalty for incorrectly predicting that a subject with Grade 0 OA has Grade 4 is the same as the penalty for predicting that the same subject has Grade 1 OA. Clearly the former represents a more serious error, yet this is not captured by evaluation measures that treat grades as categorical variables.  In this setup, permuting the ordering of the grades has no effect on classification performance. Moreover, the quantization of the KL grades to discrete integer levels is essentially an artifact of convenience; the true progression of the disease in nature is continuous, not discrete.

We therefore propose that it is more appropriate to measure the performance of an automatic knee OA severity assessment system using a continuous evaluation metric like mean squared error. Such a metric appropriately penalizes errors in proportion to their distance from the ground truth, rather than treating all errors equally. Directly optimizing mean squared error on a training set also naturally leads to the formulation of knee OA assessment as a standard regression problem. Treating it as such provides the model with more information on the structure and relationship between training examples with successive KL grades. We demonstrate that this reduces both the mean squared error and improves the multi-class classification accuracy of the model.

We fine-tuned the pre-trained BVLC CaffeNet model using both classification loss (cross entropy on softmax outputs) and regression loss (mean squared error) to compare their performance in assessing knee OA severity. In both cases, we replace fc7 with a randomly initialized layer and fine tune for 20 epochs, selecting the model with the highest validation performance. The classification network uses a 5D fully connected layer and softmax following the fc7 layer, and the regression network uses a 1D fully connected node with a linear activation. 

We compare the models using both mean squared error (MSE) and standard multi-class classification metrics. We calculated the mean squared error using the standard formula:
\begin{equation}
 MSE = \frac{1}{n} \sum_{i=1}^{n}(y_{i} - \hat{y_{i}})^{2},
\end{equation}
where $n$ is the number of test samples, $y_{i}$ is the true (integer) label and $\hat{y_{i}}$ is the predicted label. For the classification network the predicted labels $y_{i}$ are integers and for the regression network they are real numbers. We also test a configuration where we round the real outputs from the regression network to produce integer labels. Table~\ref{Tab:MSE} shows the MSE for classification using the Wndchrm and the CNN trained with classification loss (CNN-Clsf), regression loss (CNN-Reg), and regression loss with rounding (CNN-Reg*). Regression loss clearly achieves significantly lower mean squared error than both the CNN classification network and the Wndchrm features.


\begin{table}[t]
\caption{MSE for classification and regression. }
\label{Tab:MSE}
\centering
\begin{tabular}{c c c c c c}
\toprule
Classes & Wndchrm & CNN-Clsf & CNN-Reg & CNN-Reg*\\
\midrule
 Grade 0 to 4  & 2.459  & 0.836 & \textbf{0.504} & 0.576  \\
\bottomrule
\end{tabular}
\end{table}

\begin{table}[b]
\caption{
	Comparison of classification performance using classification (left) and regression (right) losses. 
}
\label{Tab:Clsf_stats}
\centering
\begin{tabular}{c c c c c c c c c c}
\toprule
      & \multicolumn{3}{c}{Classification loss} 
      & \multicolumn{3}{c}{Regression loss} \\
      \cmidrule{2-4} \cmidrule{5-7}
Grade & Precision & Recall & $F_{1}$ & Precision & Recall & $F_{1}$\\
\midrule
 0 & 0.53 & 0.64 & 0.58   & 0.57 & 0.92 & 0.71 \\
 1 & 0.25 & 0.19 & 0.22   & 0.32 & 0.14 & 0.20 \\
 2 & 0.44 & 0.32 & 0.37   & 0.71 & 0.46 & 0.56 \\
 3 & 0.37 & 0.47 & 0.41   & 0.78 & 0.73 & 0.76 \\
 4 & 0.56 & 0.54 & 0.55   & 0.89 & 0.73 & 0.80 \\
 \midrule
 Mean & 0.43 & 0.44 & 0.43 & 0.61 & 0.62 & 0.59\\
 \bottomrule
\end{tabular}
\vspace{-0.2cm}
\end{table}

To demonstrate that the regression loss also produces better classification accuracy, we compare the classification accuracy from the network trained with classification loss and the network trained with regression loss and rounded labels. Rounding, in this case, is necessary to allow for using standard classification metrics. Table~\ref{Tab:Clsf_stats} compares the resulting precision, recall, and $F_{1}$ scores. The multi-class (grade 0--4) classification accuracy of the network fine-tuned with regression loss is 59.6\%. The network trained using regression loss clearly gives superior classification performance. We suspect this is due to the fact that using regression loss gives the network more information about the ordinal relationship between the KL grades, allowing it to converge on parameters that better generalize to unseen data.


\section{Conclusion and Future Work}
This paper investigated several new methods for automatic quantification of knee OA severity using CNNs. The first step in the process is to detect the knee joint region. We propose training a linear SVM on horizontal image gradients as an alternative to template matching, which is both more accurate and faster than template matching.

Our initial approach to classifying the knee OA severity used features extracted from  pre-trained CNNs. We investigated three pre-trained networks and found that the BVLC reference CaffeNet and VGG-M-128 networks perform best. A linear SVM trained on features from these networks achieved significantly higher classification accuracy in comparison to the previous state-of-the-art. The features from pooling and convolutional layers were found to be more accurate than the fully connected layers. Fine-tuning the networks by replacing the top fully connected layer gave further improvements in multi-class classification accuracy.

Previous studies have assessed their algorithms using binary and multi-class classification metrics. We propose that it is more suitable to treat KL grades as a continuous variable and assess accuracy using mean squared error. This approach allows the model to be trained using regression loss so that errors are penalized in proportion to their severity, producing more accurate predictions. This approach also has the nice property that the predictions can fall between grades, which aligns with a continuous disease progression.

Future work will focus on improving knee joint detection accuracy using a CNN or region-based CNN instead of the proposed linear model on Sobel gradients, and on further improving assessment of knee OA severity. It is clear that the distribution of images in ImageNet and those of knee radiographs are very different. Given a large number of training examples, it would be possible to train a model from scratch on the knee OA images, which would likely be better adapted to the domain. In the absence of a large number of labeled examples, semi-supervised approaches such a ladder networks~\cite{rasmus2015semi} may prove more effective than the domain adaptation approach used here. Currently, the detection of knee joints, feature extraction, and classification/regression are separate steps. Future work will also investigate an end-to-end deep learning system by combining these steps. 

\section*{Acknowledgment}

This publication has emanated from research conducted with the financial support of Science Foundation Ireland (SFI) under grant number SFI/12/RC/2289.

The OAI is a public-private partnership comprised of five contracts (N01-AR-2-2258; N01-AR-2-2259; N01-AR-2- 2260; N01-AR-2-2261; N01-AR-2-2262) funded by the National Institutes of Health, a branch of the Department of Health and Human Services, and conducted by the OAI Study Investigators. Private funding partners include Merck Research Laboratories; Novartis Pharmaceuticals Corporation, GlaxoSmithKline; and Pfizer, Inc. Private sector funding for the OAI is managed by the Foundation for the National Institutes of Health.

\bibliographystyle{IEEEtran}
\bibliography{Biblio.bib}

\end{document}